\documentclass[11pt,a4paper]{article}

\usepackage[T1]{fontenc}
\usepackage{amsmath,amssymb,amsfonts}
\usepackage{booktabs}
\usepackage{graphicx}
\usepackage[pass,margin=1in]{geometry}
\usepackage{array}
\usepackage{hyperref}
\usepackage{textcomp}
\usepackage{float}
\usepackage{cite}

\hypersetup{
    colorlinks=true,
    linkcolor=blue,
    citecolor=blue,
    urlcolor=blue
}

\title{\textbf{HorusEye: Language as Dynamic Attention for Emergency Visual Analysis}}
\author{\textbf{Armel Yara} \\ The Day Info}

\begin{document}

\maketitle

\begin{abstract}
We introduce HorusEye, a framework for treating language as a dynamic attention mechanism in emergency visual analysis. Our study follows a five-stage pipeline. First, we benchmark RefCOCO-Degraded, a dataset of 15,244 images derived from 3,811 base images under four conditions: clean, fog, smoke, and thermal degradation. Using four research sub-questions, we evaluated several vision-language models, including Gemini, Qwen2-VL, BLIP-2, LLaVA, and Kosmos-2, in visual grounding, language-feedback recovery, VQA health, and hallucination analysis. Our main finding is that the effectiveness of language feedback is model-dependent: Gemini achieves a 47.3\% improvement under thermal conditions through iterative feedback, whereas Qwen2-VL degrades by 5.1\% under the same protocol. We also identify the ``Thermal Paradox,'' in which cropping strategies that improve RGB performance fail catastrophically in thermal imagery. In addition, BLIP-2 exhibits increased hallucination under degradation, suggesting that it is not suitable for emergency deployment.
\end{abstract}

\section{Introduction}\label{introduction}
Emergency rescue scenarios pose substantial challenges for computer vision systems. Search-and-rescue operations often occur in visually degraded environments: fog limits visibility in maritime and mountain rescue, smoke obscures scenes during fire emergencies, and thermal cameras---while useful at night---produce images that differ fundamentally from natural photographs \cite{sakaridis2018semantic,li2019benchmarking,murphy2014disaster}. Vision-language models (VLMs) have shown strong performance in visual understanding tasks; however, they are typically trained on clean, well-lit images drawn from datasets such as COCO, Visual Genome, and large-scale web image-text corpora \cite{lin2014microsoft,li2023blip2,liu2023visual}. This raises an important question for emergency applications: how well do these models perform when visual conditions deviate substantially from their training distribution?

This study investigates whether natural language feedback can function as a dynamic attention mechanism to refine visual analysis in real time under degraded emergency conditions. By ``language as dynamic attention,'' we refer to the use of natural-language prompts to guide visual processing of a model by indicating where to look and what to prioritize. This idea is analogous to the way human responders use verbal instructions to direct attention in chaotic scenes \cite{xu2015show,anderson2018bottom}.

To address this question systematically, we organized the study into four sequential research sub-questions. The first benchmarks model performance on RefCOCO-Degraded and establishes baseline visual grounding performance under degradation \cite{yu2016modeling,deng2021transvg}. The second examines whether iterative language feedback can recover accuracy lost under degraded conditions. The third evaluates whether this attention-like mechanism generalizes to a health-related visual question answer task, specifically posture classification, using full-image versus cropped-image input \cite{antol2015vqa,goyal2017making}. The fourth assesses whether visual degradation increases the risk of hallucinations, thus characterizing not only whether models fail but also how they fail \cite{rohrbach2018object,li2023evaluating,kaul2024throne}.

This paper makes four main contributions. First, we introduce RefCOCO-Degraded, a benchmark dataset containing 15,244 images with systematic fog, smoke, and thermal degradation for emergency-oriented VLM evaluation \cite{yu2016modeling,sakaridis2018semantic,li2019benchmarking}. Second, we provide empirical evidence that language feedback as dynamic attention is model-dependent: Gemini improves by 47.3\% under thermal conditions, whereas Qwen2-VL degrades by 5.1\% under the same protocol \cite{wang2024qwen2,geminiteam2024gemini}. Third, we identify the ``Thermal Paradox,'' in which cropping strategies that improve performance on RGB images fail catastrophically in thermal imagery. Fourth, we show that BLIP-2 becomes more hallucination-prone under degradation, making it unsuitable for emergency deployment \cite{li2023blip2,rohrbach2018object}.

\begin{figure}[H]
    \centering
    \includegraphics[width=0.8\linewidth]{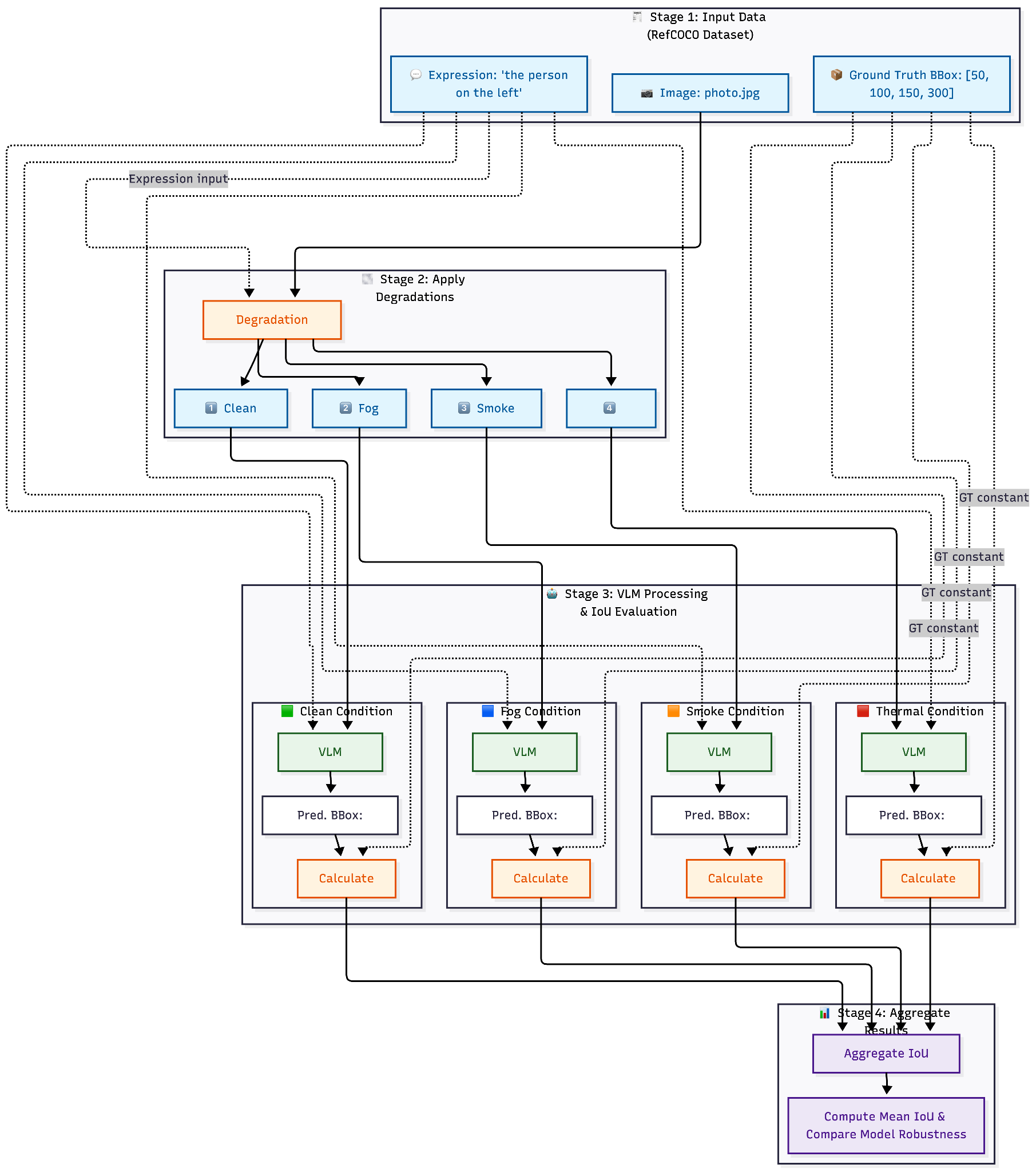}
    \caption{HorusEye Research Pipeline - Centered on the RefCOCO-Degraded dataset using iterative language feedback to dynamically adjust VLM attention for improved visual grounding and analyzes the resultant accuracy-hallucination trade-offs under emergency-relevant visual degradation.}
    \label{fig:pipeline}
\end{figure}

\section{Related Work}\label{related-work}

\subsection{Attention Mechanisms in Vision}\label{attention-mechanisms-in-vision}
The concept of attention in computer vision was formalized by Xu et al. \cite{xu2015show} in \emph{Show, Attend and Tell}, which introduced soft and hard attention mechanisms for image captioning. In that framework, the model learns to focus on relevant image regions while generating each word. Mathematically, attention computes a weighted sum over spatial locations, where the weights determine the contribution of each feature map region to the final representation. Our work extends this idea by asking whether language can serve as an external attention signal to redirect visual processing, rather than relying only on internally learned attention.

\begin{equation}
z = \sum_{i} \alpha_{i} \cdot h_{i}, \quad \text{where } \sum_{i} \alpha_{i} = 1
\end{equation}

where $\alpha_{i}$ are attention weights and $h_{i}$ are hidden states at spatial location $i$.

\subsection{Region-Based Visual Processing}\label{region-based-visual-processing}
Region-CNN, introduced by Girshick et al. \cite{girshick2014rich}, established the propose-then-classify paradigm for object detection. In this approach, the bounding box defines where to look, while cropping isolates the region for detailed analysis. Anderson et al. \cite{anderson2018bottom} combined region-based reasoning with attention in \emph{Bottom-Up and Top-Down Attention}, showing that object-level features can significantly improve VQA performance. Our health assessment experiment directly tests this paradigm by asking whether cropping, as an explicit form of spatial attention, improves health-related VQA under degradation. The answer, as shown later in the paper, is both condition-dependent and model-dependent.

\subsection{Vision-Language Models}\label{vision-language-models}
Recent VLMs have achieved impressive multimodal understanding through a range of architectural choices. BLIP-2 uses a Q-Former to connect frozen image encoders with large language models \cite{li2023blip2}. LLaVA directly projects visual features into the language embedding space \cite{liu2023visual,liu2024improved}. Gemini and Qwen2-VL rely on proprietary multimodal architectures \cite{geminiteam2024gemini,geminiteam2024gemini15,wang2024qwen2}. In this study, we systematically evaluate these diverse models under realistic emergency degradations to compare their robustness and failure modes.

\subsection{Robustness Under Distribution Shift}\label{robustness-under-distribution-shift}
Hendrycks and Dietterich \cite{hendrycks2019benchmarking} introduced ImageNet-C to benchmark neural networks under 15 corruption types. However, those corruptions, such as Gaussian noise and motion blur, differ from emergency-relevant degradations like fog, smoke, and thermal imaging. RefCOCO-Degraded was designed specifically to target the conditions encountered in search-and-rescue operations, where visibility and context are often severely compromised \cite{sakaridis2018semantic,li2019benchmarking}.

\section{Methodology}\label{methodology}

\subsection{Research Design}\label{research-design}
Our methodology follows a five-stage pipeline in which the four research sub-questions build sequentially on one another. This design allows us first to establish the problem in RSQ1, then test our central hypothesis in RSQ2, evaluate generalization in RSQ3, and assess safety in RSQ4. The overall workflow is intended to move from baseline characterization to recovery, transfer, and risk analysis.

\subsection{Models Evaluated}\label{models-evaluated}
We evaluate five vision-language models representing different architectural families: Gemini, Qwen2-VL, BLIP-2, LLaVA, and Kosmos-2 \cite{li2023blip2,liu2023visual,peng2023kosmos,wang2024qwen2}. These models were selected to capture a range of multimodal design choices, including proprietary and open-source systems, frozen-encoder frameworks, direct projection architectures, and explicit grounding-oriented models.

\begin{figure}[htbp]
    \centering
    \includegraphics[width=0.7\linewidth]{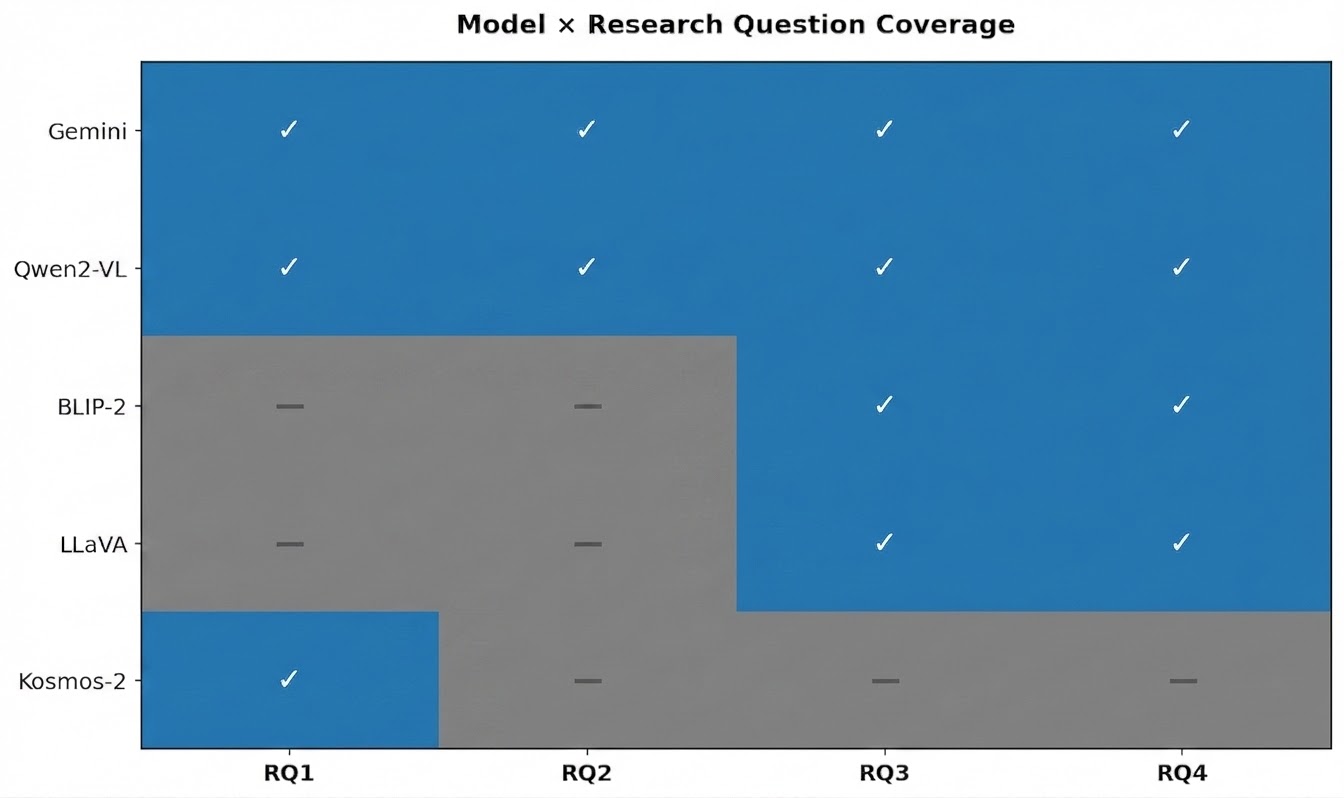}
    \caption{Model $\times$ Research Question Coverage Matrix}
    \label{fig:coverage_matrix}
\end{figure}

\begin{table}[htbp]
\centering
\caption{Vision-Language Models evaluated in this study}
\label{table:models}
\begin{tabular}{llll}
\toprule
\textbf{Model} & \textbf{Architecture} & \textbf{Parameters} & \textbf{Key Feature} \\
\midrule
Gemini 2.0 Flash & Proprietary multimodal & Not disclosed & Strongest baseline \\
Qwen2-VL-2B & Open-source VLM & 2B parameters & Instruction-tuned \\
BLIP-2 & Q-Former bridge & $\sim$3B parameters & Frozen encoders \\
LLaVA-1.6 & Visual projector & 7B parameters & Instruction-tuned \\
Kosmos-2 & Grounding model & 1.6B parameters & Explicit grounding \\
\bottomrule
\end{tabular}
\end{table}

\subsection{Evaluation Metrics}\label{evaluation-metrics}
We use task-specific metrics aligned with each research sub-question. For RSQ1 and RSQ2, performance is measured with Intersection over Union (IoU) to assess bounding-box prediction accuracy. For RSQ3, we use classification accuracy to evaluate posture recognition. For RSQ4, we define an H-Score to quantify hallucination severity, combining fabrication, overconfidence, and uncertainty-related behavior \cite{rohrbach2018object,li2023evaluating,kaul2024throne}.

\begin{equation}
\mathrm{IoU} = \frac{|B_{\text{pred}} \cap B_{\text{gt}}|}{|B_{\text{pred}} \cup B_{\text{gt}}|}
\end{equation}

\begin{equation}
\text{Accuracy} = \frac{\text{Correct Predictions}}{\text{Total Samples}} \times 100\%
\end{equation}

\begin{equation}
\text{H-Score} = \text{Fabrication} + (\text{Overconfidence} \times 0.5) - (\text{Uncertainty} \times 0.3)
\end{equation}

\section{Dataset: RefCOCO-Degraded}\label{dataset-refcoco-degraded}

\subsection{Base Dataset}\label{base-dataset}
RefCOCO serves as the base dataset for our benchmark \cite{yu2016modeling}. It contains natural images from COCO paired with human-annotated referring expressions and ground-truth bounding boxes \cite{lin2014microsoft,kazemzadeh2014referitgame}. We selected 3,811 images containing clear person references, ensuring that each example included at least one unambiguous person annotation with corresponding bounding-box ground truth.

\subsection{Degradation Conditions}\label{degradation-conditions}
Each base image is systematically transformed into four conditions designed to simulate realistic emergency scenarios: clean, fog, smoke, and thermal. The clean condition serves as the unmodified baseline. Fog is simulated using a physically inspired atmospheric scattering model, smoke is generated using particle-based occlusion with Perlin noise, and thermal imaging is approximated by converting RGB images to grayscale intensity \cite{sakaridis2018semantic,li2019benchmarking}.

\begin{figure}[htbp]
    \centering
    \includegraphics[width=0.9\linewidth]{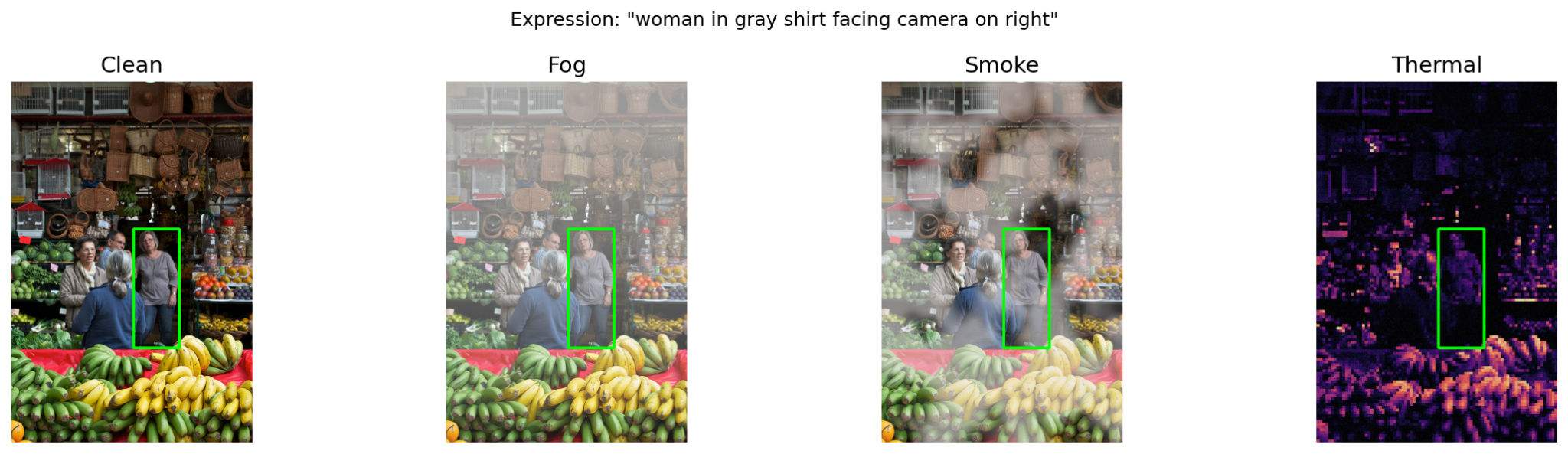}
    \caption{Visual degradation conditions in RefCOCO-Degraded}
    \label{fig:degradations}
\end{figure}

\begin{equation}
I_{\text{fog}}(x) = I(x) \cdot t(x) + A \cdot (1 - t(x)), \quad \text{where } t(x) = \exp(-\beta \cdot d(x))
\end{equation}

where $I(x)$ is the original intensity, $t(x)$ is the transmission map, $A$ is atmospheric light, and $\beta$ is the scattering coefficient. This reduces contrast while preserving color information.

\begin{equation}
I_{\text{smoke}}(x) = I(x) \cdot (1 - \alpha(x)) + C_{\text{smoke}} \cdot \alpha(x)
\end{equation}

where $\alpha(x)$ is spatially-varying opacity from noise functions. This introduces local variations and color shifts toward gray.

\begin{equation}
I_{\text{thermal}} = 0.299R + 0.587G + 0.114B
\end{equation}

\subsection{Dataset Statistics}\label{dataset-statistics}
RefCOCO-Degraded contains 15,244 images in total, corresponding to 3,811 base images across four conditions. Fog, smoke, and thermal degradation are applied at an intensity level of 0.5.

\begin{table}[htbp]
\centering
\caption{RefCOCO-Degraded dataset statistics}
\label{table:dataset_stats}
\begin{tabular}{lll}
\toprule
\textbf{Metric} & \textbf{Value} & \textbf{Notes} \\
\midrule
Base images & 3,811 & From RefCOCO \\
Conditions & 4 & Clean, Fog, Smoke, Thermal \\
Total images & 15,244 & 3,811 $\times$ 4 \\
Degradation intensity & 0.5 & For fog, smoke and thermal \\
Person annotations & 1+ per image & With referring expressions \\
\bottomrule
\end{tabular}
\end{table}

\section{Experiments}\label{experiments}

\subsection{Visual Grounding Under Degradation (RSQ1)}\label{visual-grounding-under-degradation}
We first ask how visual grounding performance changes under emergency-relevant degradation. This question establishes the baseline: before evaluating whether language feedback can help, we must quantify the extent to which performance deteriorates under adverse visual conditions.

\subsection{Method}\label{method-rsq1}
For each image, the model receives a referring expression and must predict the bounding box coordinates of the referenced person. We evaluate performance using IoU and report the mean IoU across all samples in each condition. Degradation impact is measured as the relative change from the clean baseline.

\begin{equation}
\Delta_{\text{condition}} = \frac{\text{IoU}_{\text{condition}} - \text{IoU}_{\text{clean}}}{\text{IoU}_{\text{clean}}} \times 100\%
\end{equation}

\subsection{Results}\label{rsq1-results}
The results show that thermal imagery causes the most severe degradation. Gemini drops from 0.638 on clean images to 0.397 in thermal conditions, corresponding to a 37.7\% decrease. Qwen2-VL is affected even more strongly, falling from 0.589 to 0.214, a 63.6\% decrease. Fog and smoke produce more moderate declines, suggesting that models can partially cope with reduced visibility when color information is still preserved. Kosmos-2 performs poorly even on clean images, with an IoU of approximately 0.16, making it less informative for degradation analysis.

\begin{table}[htbp]
\centering
\caption{RSQ1 Results - Mean IoU by degradation condition}
\label{table:RSQ1_results}
\begin{tabular}{lccccc}
\toprule
\textbf{Model} & \textbf{Clean} & \textbf{Fog} & \textbf{Smoke} & \textbf{Thermal} & \textbf{$\Delta$ Thermal} \\
\midrule
Gemini & 0.638 & 0.573 & 0.568 & 0.397 & -37.7\% \\
Qwen2-VL & 0.589 & 0.575 & 0.547 & 0.214 & -63.6\% \\
Kosmos-2 & 0.156 & 0.161 & 0.161 & 0.175 & +12.3\% \\
\bottomrule
\end{tabular}
\end{table}

\begin{figure}[htbp]
    \centering
    \includegraphics[width=0.8\linewidth]{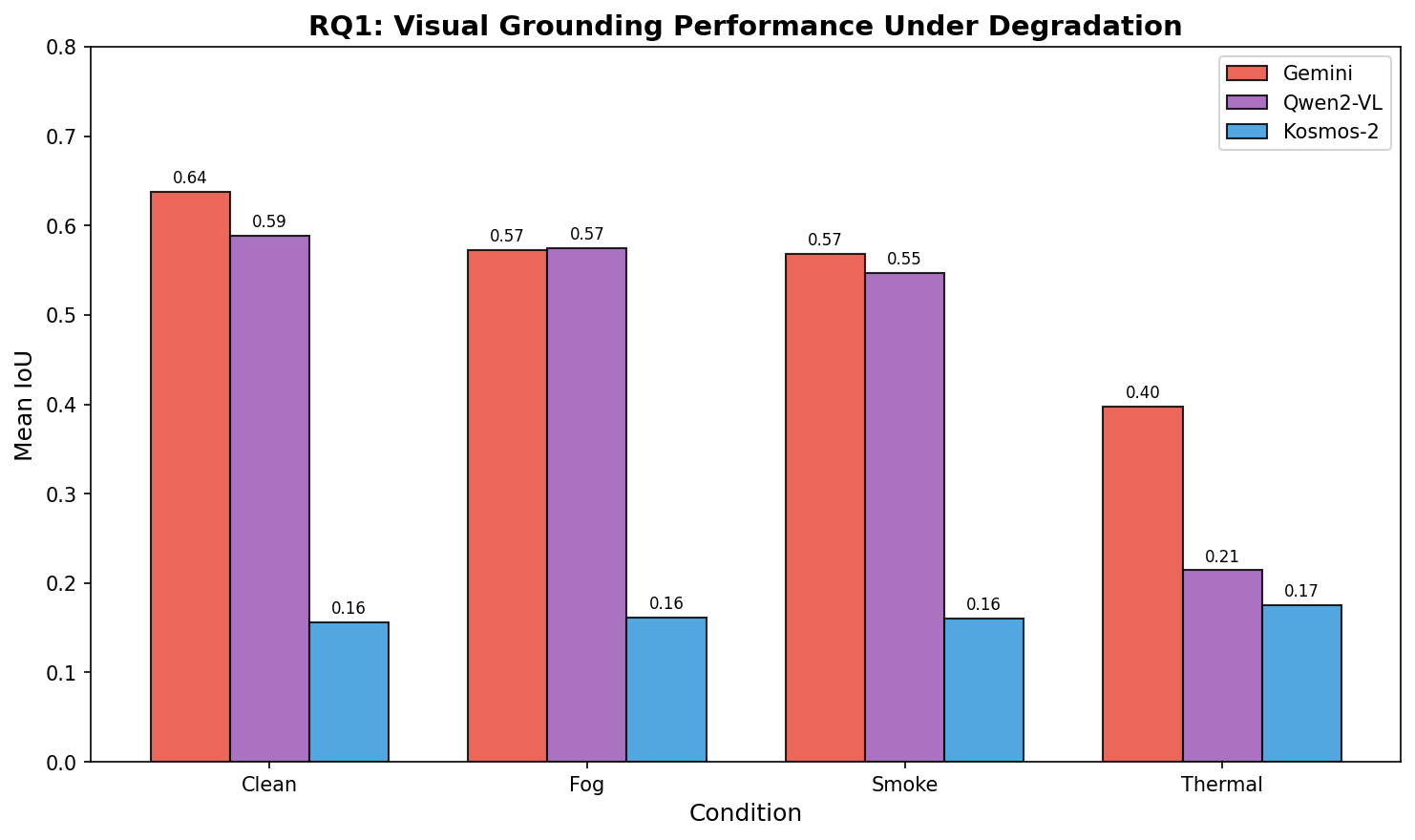}
    \caption{Visual grounding IoU comparison across conditions}
    \label{fig:grounding_bars}
\end{figure}

\subsection{Analysis}\label{analysis}
The thermal condition causes catastrophic degradation: both Gemini (-37.7\%) and Qwen2-VL (-63.6\%) show severe performance drops. Qwen2-VL loses nearly two-thirds of its grounding accuracy. In addition, fog and smoke conditions cause moderate degradation: both conditions reduce IoU by approximately 10--15\%, suggesting models can partially cope with reduced visibility when color information is preserved. Finally, the Kosmos-2 baseline is too low: with $\text{IoU} \approx 0.16$ even on clean images, Kosmos-2 struggles with the grounding task itself, making degradation analysis less meaningful.

Thermal imagery represents the most challenging condition, with Qwen2-VL losing 63.6\% of grounding accuracy. This establishes the severity of the problem that RSQ2 attempts to address.

\subsection{Language Feedback as Dynamic Attention (RSQ2)}\label{language-feedback-as-dynamic-attention}
We next evaluate whether iterative language feedback can recover grounding accuracy lost under degradation. This experiment directly tests the central hypothesis that natural language can act as a dynamic attention mechanism by redirecting the model’s visual processing.

\subsubsection{Method}\label{rsq2-method}
We use a three-round feedback protocol. In Round 1, the model makes an initial prediction using a standard referring expression prompt. In Round 2, it receives corrective feedback based on the error direction, such as ``look more to the left'' or ``look higher.'' In Round 3, the model receives a further refinement based on the second prediction. We include only cases where the initial prediction is incorrect, ensuring that the analysis focuses on recovery rather than refinement of already correct outputs.

\begin{figure}[htbp]
    \centering
    \includegraphics[width=0.9\linewidth]{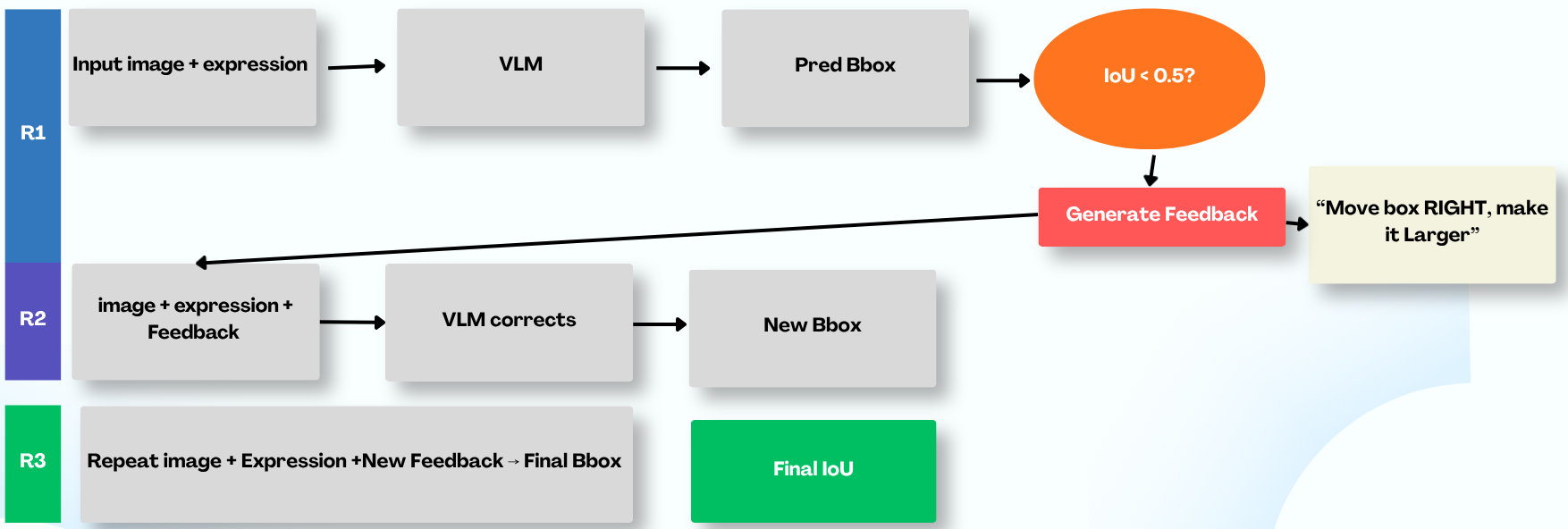}
    \caption{RSQ2 Language Feedback Pipeline}
    \label{fig:feedback_pipeline}
\end{figure}

For the filter criterion, we only process samples where R1 $\text{IoU} < 0.5$ (model initially failed). This ensures we are testing recovery, not refinement of already-correct predictions. The improvement metric is:
\begin{equation}
\text{Gain} = \frac{\text{IoU}_{\text{R3}} - \text{IoU}_{\text{R1}}}{\text{IoU}_{\text{R1}}} \times 100\%
\end{equation}

\subsubsection{Results}\label{rsq2-results}
\begin{table}[htbp]
\centering
\caption{RSQ2 Results for Gemini - IoU progression over feedback rounds}
\label{table:RSQ2_gemini}
\begin{tabular}{lcccc}
\toprule
\textbf{Condition} & \textbf{R1} & \textbf{R2} & \textbf{R3} & \textbf{Gain} \\
\midrule
Clean   & 0.218 & 0.215 & 0.219 & +0.5\%  \\
Fog     & 0.227 & 0.232 & 0.240 & +5.7\%  \\
Smoke   & 0.224 & 0.241 & 0.263 & +17.6\% \\
Thermal & 0.202 & 0.248 & 0.298 & +47.3\% \\
\bottomrule
\end{tabular}
\end{table}

\begin{table}[htbp]
\centering
\caption{RSQ2 Results for Qwen2-VL - IoU progression over feedback rounds}
\label{table:RSQ2_qwen}
\begin{tabular}{lcccc}
\toprule
\textbf{Condition} & \textbf{R1} & \textbf{R2} & \textbf{R3} & \textbf{Gain} \\
\midrule
Clean   & 0.194 & 0.212 & 0.230 & +18.7\% \\
Fog     & 0.186 & 0.201 & 0.217 & +17.0\% \\
Smoke   & 0.193 & 0.198 & 0.203 & +5.3\%  \\
Thermal & 0.115 & 0.112 & 0.109 & -5.1\%  \\
\bottomrule
\end{tabular}
\end{table}

\begin{figure}[htbp]
    \centering
    \includegraphics[width=0.8\linewidth]{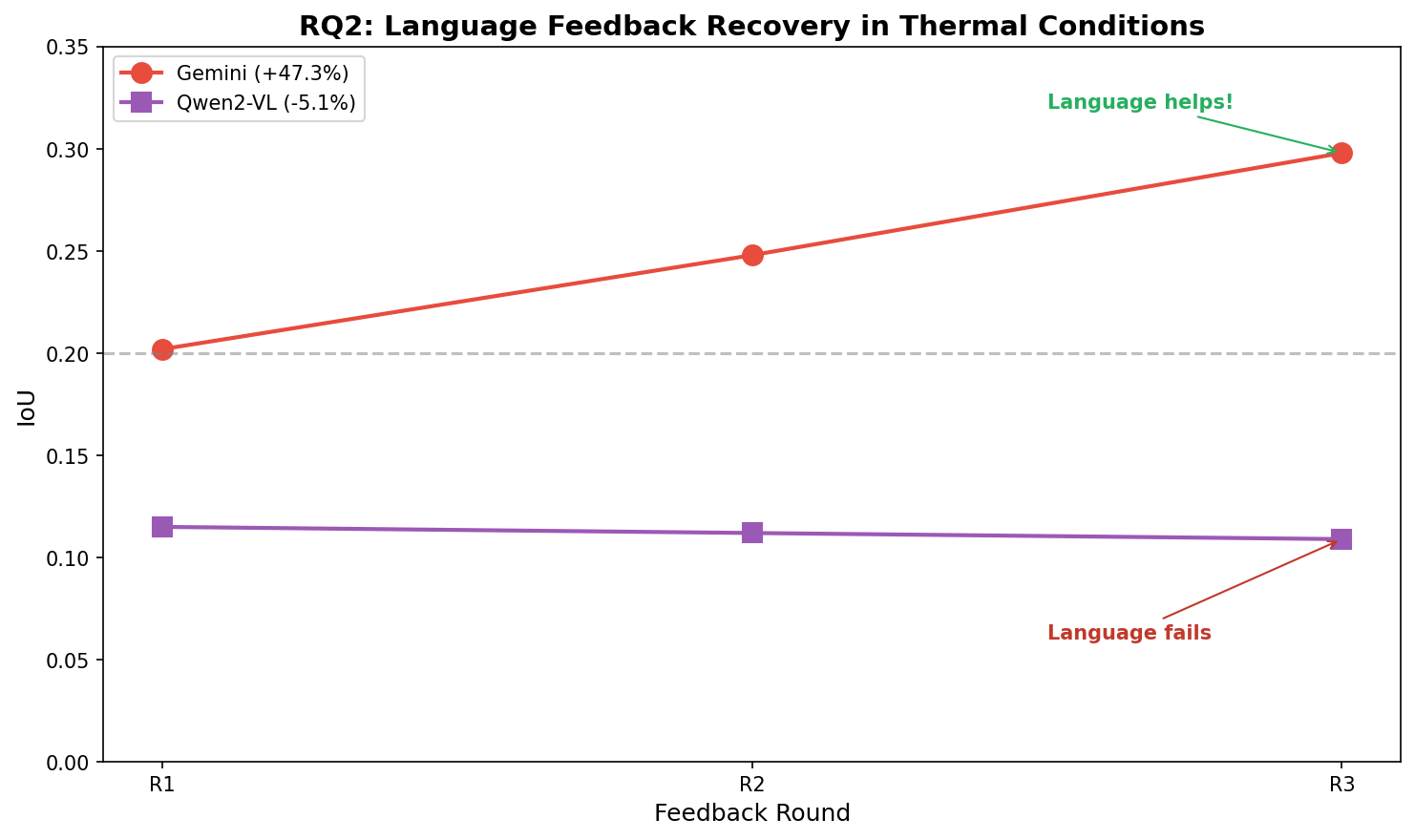}
    \caption{Thermal condition divergence - Gemini improves while Qwen2-VL degrades}
    \label{fig:thermal_divergence}
\end{figure}

\subsubsection{Analysis}\label{rsq2-analysis}
The results reveal a strong model-dependent effect. Gemini benefits substantially from iterative feedback, especially in thermal conditions, where performance improves by 47.3\%. This suggests that the model can use linguistic cues to reorient visual attention effectively. By contrast, Qwen2-VL shows a different pattern: although it improves in clean, foggy, and smoky conditions, it degrades by 5.1\% in thermal imagery. This divergence demonstrates that language-feedback effectiveness is not universal and depends strongly on the model architecture and training regime.

\subsection{Health VQA Under Degradation (RSQ3)}\label{health-vqa-under-degradation}
We then test whether explicit spatial attention through cropping generalizes to a health-related visual question answering task. Specifically, we evaluate posture classification in the categories standing, sitting, and lying, which is relevant to emergency triage.

\subsubsection{Method}\label{rsq3-method}
We evaluate posture classification (STANDING, SITTING, LYING), a critical assessment for emergency triage. Two conditions are compared:

\begin{figure}[htbp]
    \centering
    \includegraphics[width=0.9\linewidth]{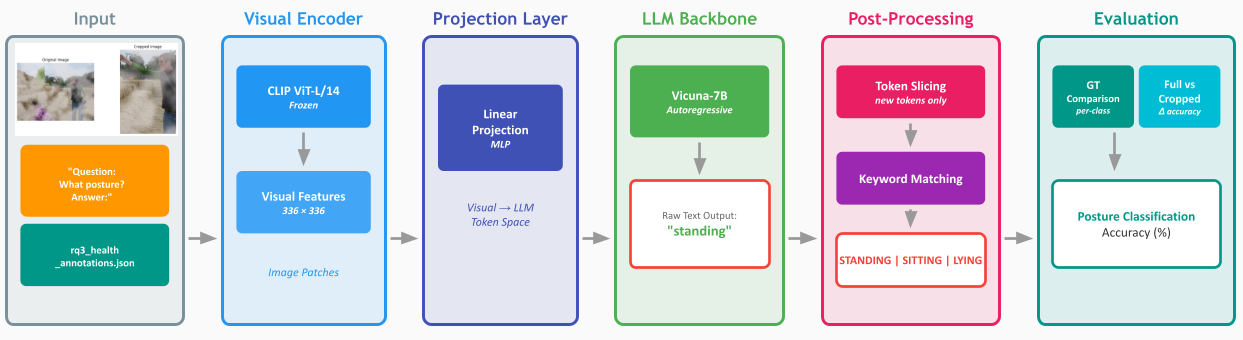}
    \caption{RSQ3 Cropping Strategy Pipeline}
    \label{fig:cropping_pipeline}
\end{figure}

The full image condition, where the model sees the entire scene and must identify the target person's posture based on the referring expression, and the cropped image condition, where the model sees only the person region extracted using the ground truth bounding box. Cropping serves as explicit attention: the bounding box defines ``where'' to look, and the crop extracts only that region. This tests whether focused attention improves or harms health assessment.

\subsubsection{Results}\label{rsq3-results}
The results show that cropping has different effects depending on the model and visual condition. For Gemini, cropping improves RGB performance but severely harms thermal performance, producing the ``Thermal Paradox.'' For Qwen2-VL, cropping generally hurts performance in RGB conditions but slightly improves it in thermal imagery. BLIP-2 benefits consistently from cropping across all conditions, while LLaVA remains largely unchanged. These results indicate that there is no universal cropping strategy that works across models and modalities.

\begin{table}[htbp]
\centering
\caption{RSQ3 Results - Posture classification accuracy}
\label{table:RSQ3_results}
\begin{tabular}{lcccccc}
\toprule
\textbf{Model} & \textbf{Cln Full} & \textbf{Cln Crop} & \textbf{$\Delta$ Cln} & \textbf{Thm Full} & \textbf{Thm Crop} & \textbf{$\Delta$ Thm} \\
\midrule
Gemini & 94\% & 98\% & +4\% & 94\% & 68\% & -26\% \\
Qwen2-VL & 96\% & 84\% & -12\% & 68\% & 70\% & +2\% \\
BLIP-2 & 76\% & 92\% & +16\% & 58\% & 70\% & +12\% \\
LLaVA & 64\% & 64\% & 0\% & 64\% & 64\% & 0\% \\
\bottomrule
\end{tabular}
\end{table}

\begin{figure}[htbp]
    \centering
    \includegraphics[width=0.9\linewidth]{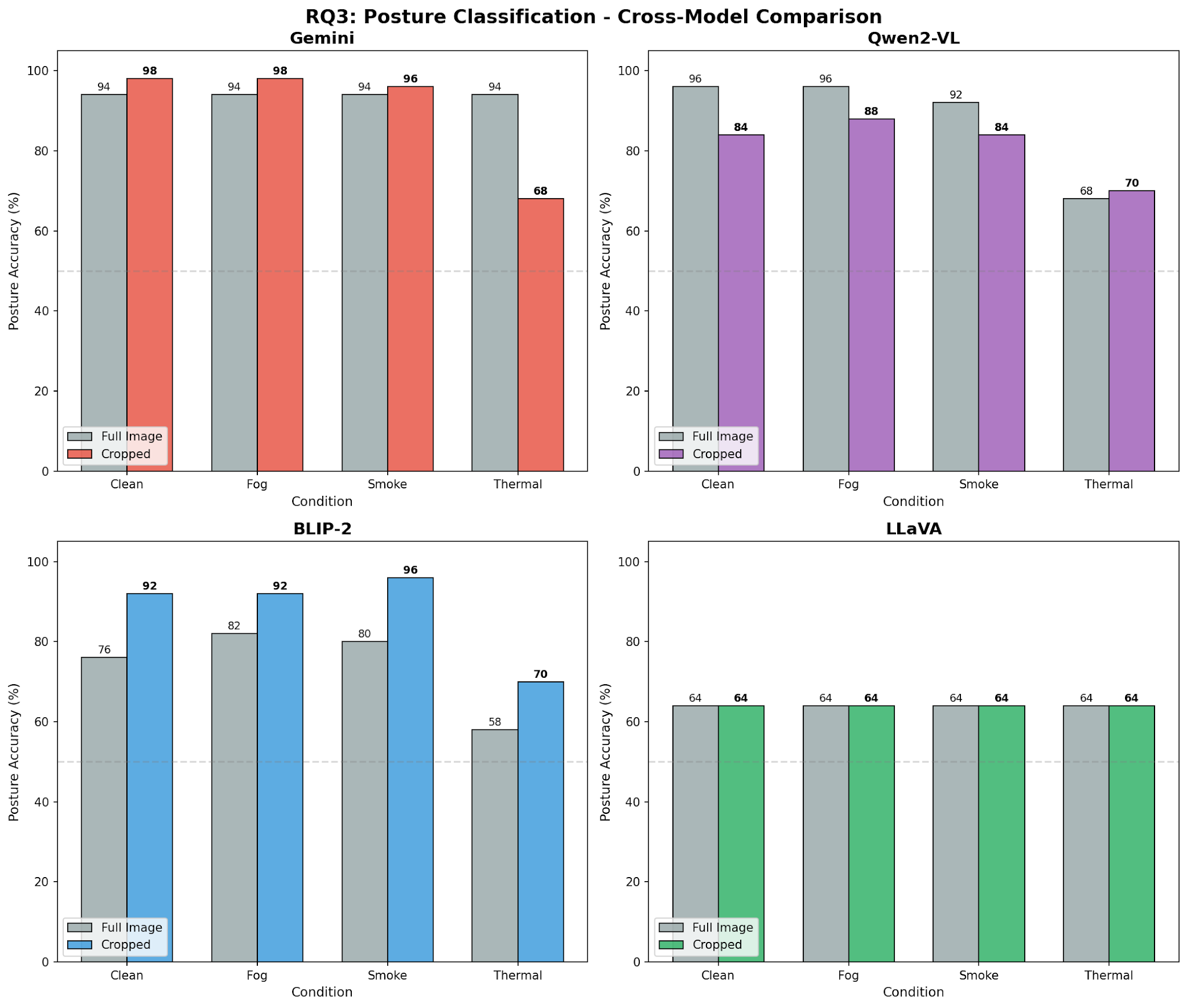}
    \caption{Cross-model accuracy comparison for full vs cropped images}
    \label{fig:cross_model_accuracy}
\end{figure}

\subsubsection{The Thermal Paradox}\label{the-thermal-paradox}
We discover a critical phenomenon we term the Thermal Paradox, in which cropping strategies that improve performance in RGB conditions (+4\% for Gemini) cause catastrophic failure in thermal conditions (-26\%). The same technique has opposite effects depending on image modality.

This finding reveals that in thermal imagery, cropping removes crucial contextual cues needed for posture inference---such as body orientation relative to the ground/horizon, surrounding objects like chairs or beds, and spatial relationships within the scene. When color and texture are absent (thermal), these contextual cues become essential for posture classification.

\subsubsection{Model Behavior Summary}\label{model-behavior-summary}
\textbf{BLIP-2} is most robust to cropping. It benefits from focused attention in all conditions (+12\% to +16\%). The Q-Former architecture appears to handle reduced context well. \textbf{Qwen2-VL} requires full scene context. Cropping consistently hurts RGB performance (-8\% to -12\%). \textbf{Gemini} is condition-dependent. Cropping helps RGB but destroys thermal. \textbf{LLaVA} is ``insensitive.'' It shows no response to cropping or degradation (constant 64\%), suggesting possible mode collapse to the majority class.

\subsection{Hallucination Under Degradation (RSQ4)}\label{hallucination-under-degradation}
Finally, we examine whether visual degradation increases hallucination risk. This is critical for emergency deployment, because a model that confidently produces incorrect information may be more dangerous than one that expresses uncertainty.

\subsubsection{Method}\label{rsq4-method}
We define an H-Score to capture hallucination severity by combining fabrication, overconfidence, and uncertainty-related behavior. Higher scores indicate more dangerous hallucination, while lower or negative scores indicate more appropriate uncertainty.

\begin{figure}[htbp]
    \centering
    \includegraphics[width=0.9\linewidth]{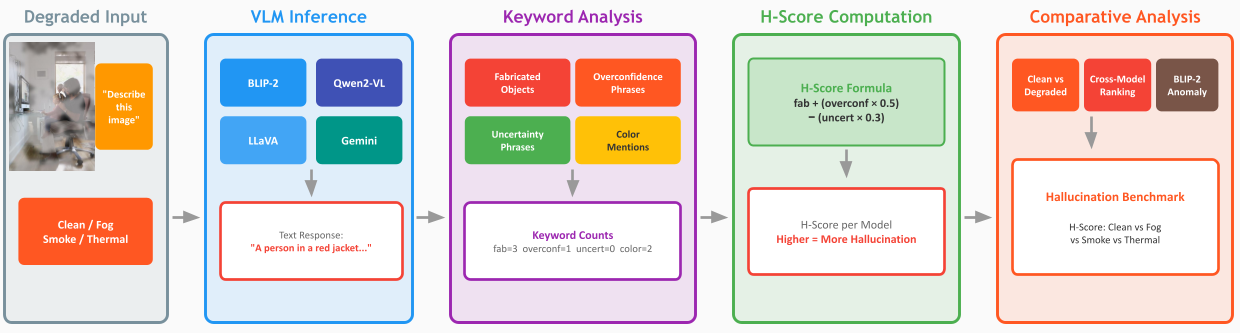}
    \caption{H-Score formula and components}
    \label{fig:hscore_components}
\end{figure}

\begin{equation}
\text{H-Score} = \text{Fabrication} + (\text{Overconfidence} \times 0.5) - (\text{Uncertainty} \times 0.3)
\end{equation}

We define the components as follows. Fabrication is the count of objects, attributes, or details mentioned that do not exist in the image, detected via keyword matching against known fabrication indicators. Overconfidence refers to the use of definitive language such as ``definitely,'' ``clearly,'' or ``obviously'' when the visual evidence is ambiguous or degraded. Uncertainty captures appropriate hedging language such as ``appears to be,'' ``might be,'' or ``it's difficult to tell'' when visibility is genuinely limited. Higher H-Score means more dangerous hallucination; therefore, a negative H-Score indicates that the model appropriately expresses uncertainty.

\subsubsection{Results}\label{rsq4-results}
\begin{table}[htbp]
\centering
\caption{RSQ4 Results - H-Score by degradation condition}
\label{table:RSQ4_results}
\begin{tabular}{lcccccc}
\toprule
\textbf{Model} & \textbf{Clean} & \textbf{Fog} & \textbf{Smoke} & \textbf{Thermal} & \textbf{$\Delta$ (C$\rightarrow$T)} & \textbf{Status} \\
\midrule
Gemini & 0.44 & 0.43 & 0.44 & -0.05 & -0.49 & SAFER \\
Qwen2-VL & 0.17 & 0.19 & 0.13 & -0.24 & -0.41 & SAFER \\
BLIP-2 & 0.35 & 0.36 & 0.49 & 0.69 & +0.34 & DANGER \\
LLaVA & 1.69 & 1.70 & 1.73 & 1.40 & -0.29 & $\sim$ Stable \\
\bottomrule
\end{tabular}
\end{table}

\begin{figure}[htbp]
    \centering
    \includegraphics[width=0.8\linewidth]{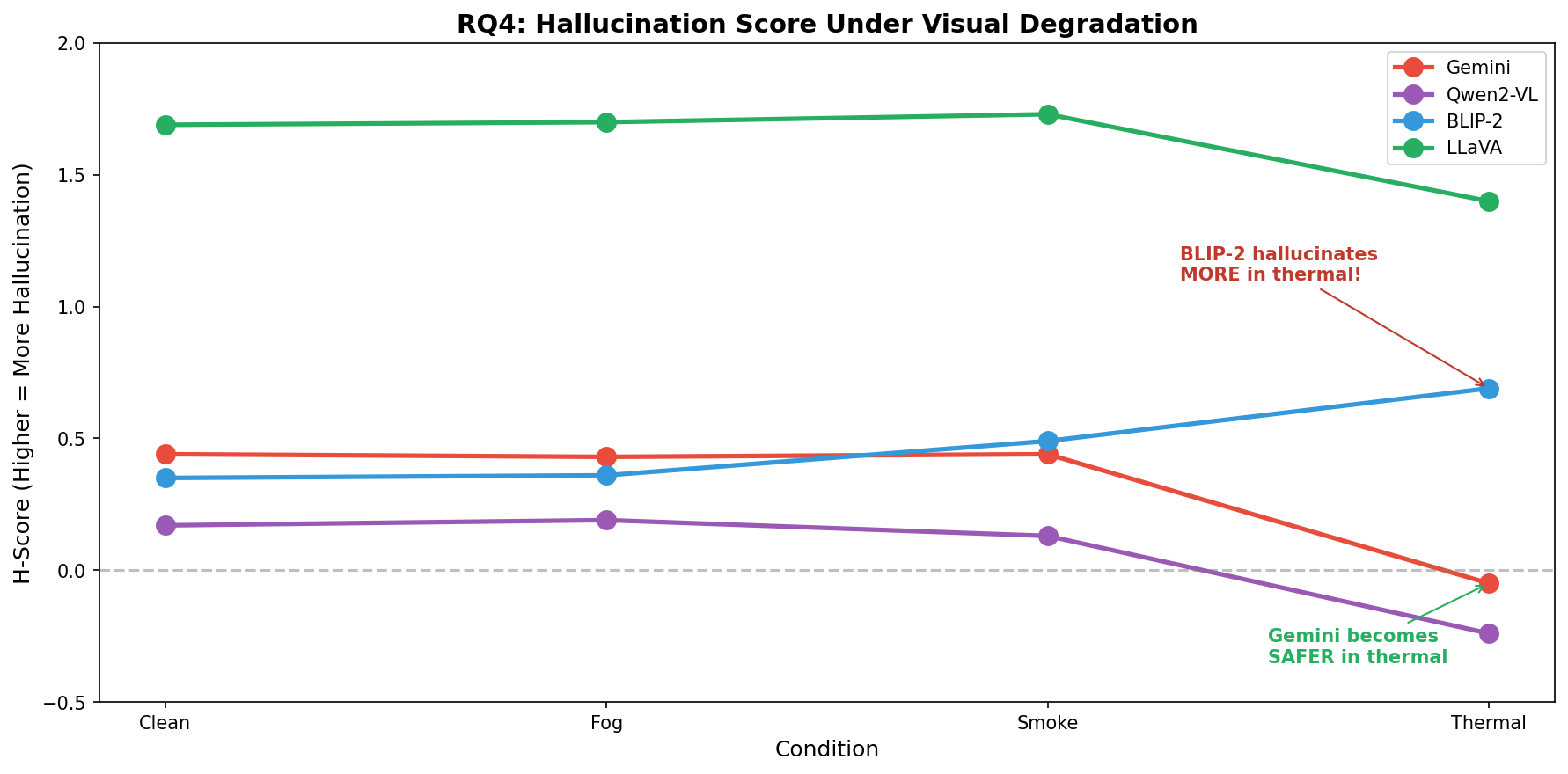}
    \caption{H-Score comparison across models and conditions}
    \label{fig:hscore_comparison}
\end{figure}

\section{Discussion}\label{discussion}
\subsection{Answer to the Main Research Question}
The central question of this study was whether natural language feedback can function as a dynamic attention mechanism under degraded emergency conditions. Our results indicate that the answer is yes, but only in a model-dependent manner. Gemini demonstrates clear gains from iterative feedback in the most challenging thermal setting, suggesting that linguistic cues can successfully redirect visual attention and help recover degraded performance. In contrast, Qwen2-VL does not benefit uniformly from the same protocol and even degrades in thermal imagery, showing that language-guided attention is not a universal property of vision-language models.

\subsection{Key Findings Summary}
Several important patterns emerge from the experiments. First, thermal imagery is consistently the most difficult condition for grounding, confirming that modality shift creates a substantial challenge for models trained primarily on clean natural images. Second, the effectiveness of language feedback increases as visual degradation worsens for some models, particularly Gemini, which suggests that corrective linguistic prompts may become more valuable when visual evidence is sparse. Third, cropping improves performance only under certain conditions, and in thermal imagery it may remove essential contextual information needed for correct posture inference. This is the basis of the Thermal Paradox: spatial focus can help in RGB settings but harm performance when the scene depends heavily on global context.
In summary, RSQ1 shows that thermal imagery is the most challenging condition, with a -63.6\% IoU loss for Qwen2-VL. RSQ2 reveals that language feedback has model-specific effects (+47.3\% for Gemini vs -5.1\% for Qwen). RSQ3 indicates that cropping improves RGB performance but reduces thermal IoU, so no universal cropping strategy exists. RSQ4 demonstrates that BLIP-2 hallucinates more under degradation, making it unsafe for emergency deployment.
A major result of this study is that hallucination behavior also changes under degradation. BLIP-2 is the most concerning model in this regard, because it becomes more hallucination-prone as visual quality decreases and does not reliably express uncertainty. In a search-and-rescue context, this behavior is especially dangerous, since a model that confidently invents objects or attributes may mislead responders. By comparison, Gemini and Qwen2-VL generally respond to degradation with more uncertainty, which is safer than overconfident fabrication.

\subsection{Practical Recommendations}\label{practical-recommendations}
These findings have direct implications for emergency AI deployment. First, model selection matters: performance under clean conditions is not sufficient to guarantee robustness in degraded environments. Second, attention strategies such as cropping or language feedback must be validated for each model and condition rather than assumed to transfer across architectures. Third, systems intended for emergency use should prefer models that degrade gracefully and express uncertainty when the input is ambiguous. In practice, this means that safety-critical applications should avoid relying on a single strategy or a single benchmark condition.

\section{Limitations}\label{limitations}
This study has several limitations. The degradations used here are synthetic, and although they are physically motivated, they may not fully capture the complexity of real emergency sensor data. In addition, the health-VQA experiments were conducted on a limited sample size, which restricts the strength of broader generalizations. The findings are also specific to the evaluated model versions and may change as newer releases become available.

\section{Future Work}\label{future-work}
Future work should focus on validation with real emergency imagery, fine-tuning VLMs on degraded data, and developing model-agnostic attention mechanisms that can operate reliably across different architectures. Multi-turn prompting strategies that adapt dynamically to model responses may also improve robustness in operational settings.

\section{Conclusion}\label{conclusion}
This study examined whether natural language feedback can serve as a dynamic attention mechanism for vision-language models under degraded emergency conditions. Using the RefCOCO-Degraded benchmark and four research questions across five models, we showed that the answer is affirmative, but strongly dependent on both model architecture and visual condition.

Our experiments demonstrate that thermal imagery is the most challenging setting for visual grounding, with substantial performance loss for all models except the weakest grounding baseline. We also found that iterative language feedback can substantially improve performance for some models, most notably Gemini, while failing or even degrading performance for others such as Qwen2-VL. This confirms that language-guided attention is not a universal capability, but rather an emergent behavior that depends on how each model encodes and uses multimodal information.

In addition, we identified the Thermal Paradox, where cropping strategies that improve performance on RGB images can fail catastrophically in thermal imagery. This finding shows that explicit spatial focus is not always beneficial and that contextual information may be essential in low-appearance modalities. Finally, we found that BLIP-2 becomes more hallucination-prone under degradation, making it a risky choice for emergency deployment unless hallucination mitigation is applied.

Overall, our results suggest that emergency-facing vision-language systems must be evaluated under the specific visual conditions in which they will be deployed. Robustness, uncertainty calibration, and model-specific attention behavior should be treated as essential design criteria rather than optional refinements. The RefCOCO-Degraded benchmark provides a foundation for future work on safer and more adaptive multimodal systems in high-stakes environments.

\bibliography{references}

\end{document}